%% file: acl_latex.tex
\documentclass[11pt]{article}

\usepackage[preprint]{acl}

\usepackage{times}
\usepackage{latexsym}

\usepackage[T1]{fontenc}

\usepackage[utf8]{inputenc}

\usepackage{microtype}

\usepackage{inconsolata}

\definecolor{ForestGreen}{rgb}{0.28,0.6,0.02}
\definecolor{CustomBlue}{rgb}{0.125,0.29,0.53}

\usepackage{graphicx}
\usepackage{subcaption}
\usepackage{mwe} %
\usepackage[inline]{enumitem}

\usepackage{amsmath}
\usepackage{amssymb}
\usepackage{amsfonts}

\usepackage{booktabs}
\usepackage{multirow}

\usepackage{todonotes}

\usepackage[ruled,vlined]{algorithm2e}

\DeclareMathOperator*{\argmax}{arg\,max}
\DeclareMathOperator*{\argmin}{arg\,min}

\usepackage{cleveref}

\newcommand{\circnum}[1]{{\textcircled{\raisebox{-.1pt} {\tiny\textsf #1}}}}

\title{Success and Cost Elicit Convention Formation for Efficient Communication}

\author{
 \textbf{Saujas Vaduguru\textsuperscript{1}}\;\;\;
 \textbf{Yilun Hua\textsuperscript{2}}\;\;\;
 \textbf{Yoav Artzi\textsuperscript{2}}\;\;\;
 \textbf{Daniel Fried\textsuperscript{1}}
\\
\\
 \textsuperscript{1}Carnegie Mellon University\\
 \textsuperscript{2}Department of Computer Science and Cornell Tech, Cornell University
\\
\\
\texttt{\{svadugur,dfried\}@cs.cmu.edu}\;\;\;
\texttt{\{yilunhua,yoav\}@cs.cornell.edu}
}

\begin{document}
\maketitle
\begin{abstract}
Humans leverage shared conversational context to become increasingly successful and efficient at communicating over time. One manifestation of this is the formation of \emph{ad hoc} linguistic conventions, which allow people to coordinate on short, less costly utterances that are understood using shared conversational context. We present a method to train large multimodal models to form conventions, enabling efficient communication. Our approach uses simulated reference games between models, and requires \emph{no} additional human-produced data. 
In repeated reference games involving photographs and tangram images, our method enables models to communicate efficiently with people: reducing the message length by up to 41\% while increasing success by 15\% over the course of the interaction. Human listeners respond faster when interacting with our model that forms conventions. We also show that training based on success or cost alone is insufficient --- both are necessary to elicit convention formation.
\end{abstract}

\input{sections/introduction}
\input{sections/related_work}
\input{sections/method}
\input{sections/experiments}
\input{sections/discussion}

\section*{Limitations}
We demonstrate our method only in the setting of repeated reference games, which allow us to train and evaluate models in temporally extended interactions. Additionally, since we work in the domain of reference games, the entire interaction history is available in the context window to the speaker and listener models. Extending to more complex domains would involve managing much longer contexts, that may not fit in the (effective) context window of the models. We leave the exploration of these questions for future work.

\section*{Potential risks and considerations}
A model that communicates its intent as efficiently as possible may not adhere to other ideals or values. For example, a model that makes a product recommendation may convince you to choose a specific product in a short interaction, but may do so by withholding important information about the product, hence acting deceptively. Language using agents in the real world have to tradeoff multiple values, of which communicative success and cost are only two. While we show that our method enables models to communicate more efficiently (in the benign setting of reference games), broader uses will have to consider other relevant values.

\section*{Acknowledgments}
We thank Aprameya Bharadwaj for support during the initial stages of this work, and Evan Wang for helpful early discussions. We thank Will McCarthy and Priyan Vaithilingam for helpful discussions for the human study. We also thank Mustafa Omer Gul for help with tangram data. We acknowledge a gift from Google on Action, Task, and User Journey Modeling that supported this work. We also thank Modal Labs for providing compute credits. This research was supported by the NSF under grant No. 2504533. Any opinions, findings and conclusions or recommendations expressed in this material are those of the author(s) and do not necessarily reflect the views of the National Science Foundation or the other funders.

\bibliography{custom}

\appendix
\input{sections/appendix/base-model}
\input{sections/appendix/contexts}
\input{sections/appendix/training}
\input{sections/appendix/metrics}
\input{sections/appendix/prolific}
\input{sections/appendix/analysis}
\end{document}

%% file: sections/introduction.tex
\begin{figure}[t]
    \centering
    \includegraphics[width=0.85\linewidth]{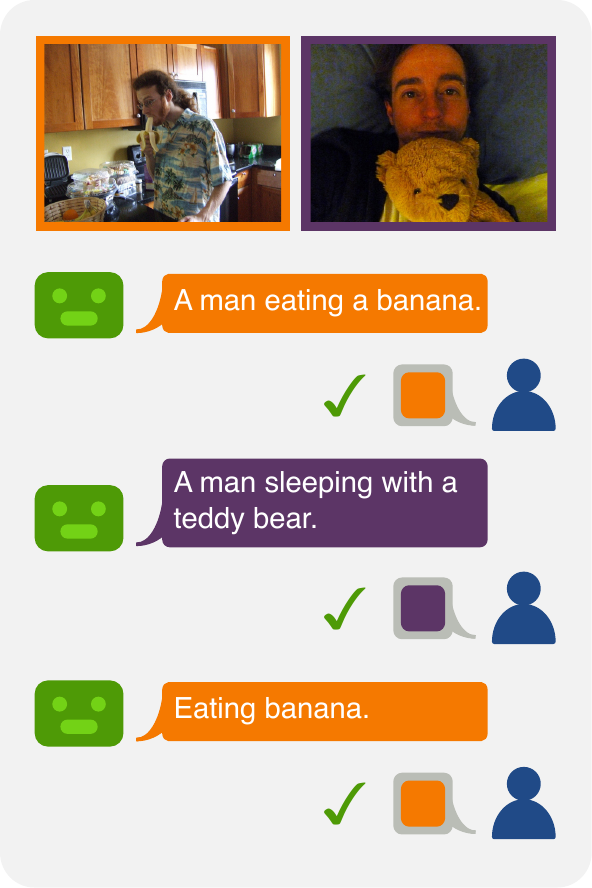}
    \caption{A real example interaction with our model playing a repeated reference game. In each turn, the \textcolor{ForestGreen}{speaker} model has to describe a single image in the context of a set of images and previous interactions. A \textcolor{CustomBlue}{listener} tries to guess the image being described. We train speaker models that adapt to communicate efficiently (using fewer words) with people by forming \emph{ad hoc} conventions.}
    \label{fig:intro-figure}
    \vspace{-2em}
\end{figure}

\section{Introduction}

As people interact with each other in natural language, they adapt to shared conversational context to communicate more efficiently. One way people adapt is by forming \emph{ad hoc} conventions, where a pair of agents enrich the standing meaning of words or phrases with information drawn from the history of their interaction. You might direct a first-time visitor to meet you \emph{``on the eighth floor patio of the computer science building''} but on the second visit, you might just agree to meet them \emph{``on the patio.''}

In this paper, we make models adapt to communicate more efficiently with people in natural language by forming conventions. We finetune large multimodal models (LMMs) to adapt in context to prior interactions with a person.
Even though our approach uses simulated games between models to produce training data and relies on \emph{no} additional human-produced data, the models we train adapt to communicate with people with increasing success and efficiency \citep{lazaridou-etal-2020-multi}.

We demonstrate our method by training models to play repeated reference games over images (\Cref{fig:intro-figure}). Repeated reference games \citep{Krauss1964ChangesIR,CLARK19861,hawkins-etal-2020-characterizing} have been used to investigate such adaptive behavior in humans. In these games, a \emph{speaker} is repeatedly directed to describe one image in the context of a set of images to a \emph{listener} that guesses the image to which the speaker is referring. Experiments with people playing games over abstract images \citep{hawkins-etal-2020-characterizing} and photographs \citep{hawkins-etal-2020-continual} reveal that people rapidly adapt to their conversational partner to use increasingly short utterances while improving communicative success.
Prior work has found that LMMs \citep{hua2024talk} and large language models (LLMs; \citeauthor{hua2025post}, \citeyear{hua2025post}) do not exhibit this behavior, and maintain similar utterance cost over the course of the interaction.

We start with pretrained speaker and listener models and simulate games between them. We apply simple notions of communicative success and cost to create preference pairs for DPO-style policy optimization \citep{dpo}. We apply our procedure to train speaker models, and evaluate them in repeated reference games over MS COCO photographs and tangram images. In games with human listeners, we find that our model reduces message length by 41\% while increasing communicative success by 16\% over the course of games over COCO images. Over the course of games over tangram images, our speaker model reduces message length by 29\% while increasing success by 27\%. We also find that human listeners adapt to respond faster to our models' utterances compared to those from a model that does not adapt. By the 4th repetition of an image, people respond 52\% faster to our COCO model's utterances, and 35\% faster to our tangram model's utterances.

We find that both components of our utility function -- communicative success and cost -- are essential. Ablations that remove one or the other do not elicit the same behavior. 
We find that while a utility function based on success alone leads to an increase in success over the interaction, it also leads to 7\% \emph{increase} in message length over the course of the interaction. On the other hand, defining utility based on cost alone hinders the model's ability to communicate successfully, even though the messages are shorter.
In contrast, our approach that defines communicative utility based on both these factors yields models that communicate with people with increasing success \emph{and efficiency}.\footnote{Our code is available at \url{https://github.com/saujasv/learning-conventions}}

%% file: sections/related_work.tex
\section{Related work}

\paragraph{Convention formation in humans and models}
Extensive prior work has shown how people adapt their language to their partners to communicate efficiently.
\citet{Krauss1964ChangesIR} find that speakers shorten their utterances in repeated interactions. \citet{CLARK19861} study the collaborative aspects of reference, including minimization of effort for both partners. \citet{hawkins-etal-2020-characterizing} collect a large-scale dataset of repeated reference games over tangrams. They characterize the changes in utterances as speakers adapt, finding that semantic meaning stabilizes within a pair of interlocutors, even as it diverges across pairs.

\citet{hua2024talk} study in-context conversational adaptation in off-the-shelf multimodal models using repeated reference games over COCO images, and formalize the notion of stable meaning within a pair using the notion of word novelty. They find that absent heavy-handed prompting, even highly capable multimodal models don't leverage context to communicate more efficiently, continuing to generate detailed descriptions even when they have diminishing communicative utility. In our work, we train models to adapt in context to communicate more efficiently over time, and elicit convention formation in multimodal models.

\input{sections/blocks/method-figure}

\paragraph{Machine learning for convention formation} 
\citet{takmaz-etal-2020-refer} present an approach to generating referring expressions by training on reference chains produced by people in the PhotoBook game \citep{haber-etal-2019-photobook}, which exhibit linguistic traits such as decreasing cost and consistent use of words in subsequent mentions.
\citet{hua2025post} propose a post-training approach for LLMs to generate references in conversation. They curate preference training data from reference chains in TV show scripts, and show that this induces convention formation when models play repeated reference games over sets of words. These approaches adapt human-produced reference chains to train models to exhibit convention formation behaviors. In contrast, the approach we propose does not rely on any additional data from extended interactions (beyond pretraining) produced by people.

\citet{hawkins-etal-2020-continual} frame conversational adaptation as domain adaptation, and propose updating model weights online during an interaction. They use syntactic structure to determine reusable parts of a referring expression, and induce convention forming behavior in models that play image reference games. In comparison, our approach learns from simulated games between models, and doesn't require updating model weights at inference time. We also create training data using simple measures of success and cost, without relying on explicit syntactic structure.

\citet{pmlr-v139-zhu21d} present an approach to adapt to a listener by modeling their mental state. They maintain an explicit neural model of the listener that is trained with meta-learning to guess the listener's response based on prior interactions. They find that this leads to utterances that are increasingly successful due to the speaker adapting to the listener. However, their approach does not focus on improving communicative efficiency. In contrast, our approach improves communicative efficiency, and does not require an explicit listener model.

\paragraph{Communicative cost in large language models} Training language models with reinforcement learning from human feedback (RLHF) has been shown to increase response length \citep{singhal2024a,dubois2024lengthcontrolled}. To inhibit length increase due to RLHF, \citet{Wu2024MetaRewardingLM} include pairs that prefer shorter responses between similarly successful ones in preference data. We use a similar mechanism to penalize message cost, and show that this leads to models that communicate increasingly efficiently with people over time.

\paragraph{Learning from simulated games}
We extend the research program proposed by \citet{lazaridou-etal-2020-multi}, who argue for using multi-agent simulations to make models trained on generic language data functional, utility-driven agents that communicate with people in natural language. We use simulations to train models to adapt to communicate efficiently.
\citet{konyushkova2025visionlanguagemodeldialoggames} present a self-play method to train multimodal models to play image identification games, and find that it improves performance on tasks like visual question answering. We also train models that play image identification games, but look at repeated games and adaptation over time rather than a single multi-turn interaction. \citet{horst2025playpenenvironmentexploringlearning} present an environment with games \citep{chalamalasetti-etal-2023-clembench} that can be used to compare algorithms for training models to interact with people.

%% file: sections/blocks/method-figure.tex
\begin{figure*}[t]
    \centering
    \includegraphics[width=0.95\linewidth]{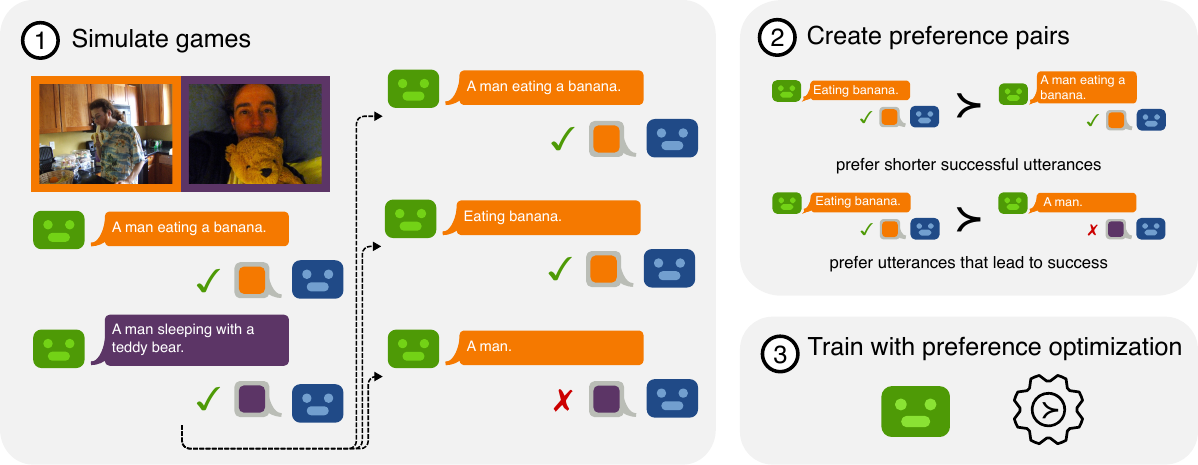}
    \caption{
        \circnum{1} We simulate interactions between \textcolor{ForestGreen}{speaker} and \textcolor{CustomBlue}{listener} models, sampling multiple descriptions for a target image at the same stage of the interaction.
        \circnum{2} We create preference pairs based on the communicative utility of descriptions. 
        \circnum{3} We train the speaker with preference optimization using the pairs created from simulated games.}
    \label{fig:method_fig}
    \vspace{-1em}
\end{figure*}

%% file: sections/method.tex
\section{Learning in simulated repeated reference games} \label{sec:method}
We consider repeated reference games defined by a context $C = \{c_1, c_2, \ldots, c_k\} \subset \mathcal{C}$ of $k$ images and a sequence of trials $t_j = (c_j, u_j, l_j)$ where $c_j \in C$ is the target, $u_j \in \mathcal{U}$ is the speaker's utterance, and $l_j \in C$ is the listener's guess in that trial. In the $i^\text{th}$ trial, the speaker observes $C$, a sequence of previous trials $T = [t_1, t_2, \ldots, t_{i-1}]$, and the next target $c_i \in C$, and produces an utterance $u_i$. The listener observes the same context $C$, the same sequence of trials $T$ and utterance $u_i$, and guesses a referent $l_i \in C$. The pair of agents is successful if $l_i = c_i$. The game proceeds in \emph{blocks} of trials, where each block includes each image being presented exactly once as the target, in a random order.

\paragraph{Simulating repeated reference games} \label{sec:simulation}
We simulate repeated reference games between a speaker model and a listener model. The speaker is modeled as a distribution $S_\theta(u|C, T_{:i}, c_i)$ with parameters $\theta$. The listener is modeled as a distribution $L_\phi(l|C, T_{:i}, u_i)$ parameterized by $\phi$. We simulate games (\Cref{alg:sim}) by choosing a context of images, and simulating a sequence of trials. We start with an empty sequence of trials $T = [\ ]$ and incrementally add trials to it. For each trial, we choose the next target $c_i$ and sample $n$ utterances $u_1, u_2, \ldots, u_n \sim S_\theta(\cdot|C, T, c_i)$ from the speaker. We present each utterance to the listener and obtain a guess $l_j = \argmax_{l} L_\phi(l|C, T, u_j)$ for $j = 1, 2, \ldots, n$, and construct a set of candidate trials $V = \{(c_i, u_j, l_j)|1 \leq j \leq n\}$. We sample a trial from $V$ to add to the running list of trials $T$, and repeat for $N$ trials.

\paragraph{Communicative utility}
Speakers may be viewed as rational actors \citep{grice,rsa} that choose an utterance according to its utility. An utterance is useful if it successfully communicates the speaker's intent. It is also useful if it is easy for the speaker to produce, or easy for the listener to comprehend. A short, easy to produce utterance might not lead to the listener inferring the speaker's intent, but a detailed utterance that would succeed might be expensive to produce. The speaker has to trade off these notions of utility. In defining utility, we consider success to be whether the listener guesses the intended referent, and cost of an utterance $u$ to be the number of words $|u|$ in the utterance \citep{Bergen2016PragmaticRT}.

\input{sections/blocks/simulation-alg}

\paragraph{Preference pairs} \label{sec:pref_pairs}
One method to optimize communicative utility would be to create a scalar utility value that can be optimized with policy gradient methods. However, trading off communicative success and cost in a single value involves searching over a large space of functions that combine these terms, with some combinations being difficult to optimize \citep{lee2019learningautocompletesystemscommunication}. We instead build on the success of learning from pairwise preferences \citep{dpo}, and create preference pairs where one sample has higher utility than the other.

For a given game state, defined by the context $C$ and the sequence of previous trials $T$, \Cref{alg:sim} gives us a set of candidate trials $V$. To create preference pairs, we consider the shortest utterance in $V$ that succeeds, $t^* = \argmin_{\{t \in V | l(t) = c(t)\}} |u(t)|$.
We then create preference pairs $t_w, t_l = (t^*, t)$ for each trial $t \in V$ where $u(t)$ is longer than $u(t^*)$, or leads to an incorrect guess
$$\{(t^*, t) | t \in V, |u(t)| > |u(t^*)| \vee l(t) \ne c(t)\}$$
We refer to this instantiation of communicative utility as \textsc{success+cost}.
This results in a dataset 
$$\mathcal{D} = \{(C, T,t_w, t_l) | t_w, t_l \in V\}$$ 
where we prefer that the speaker produce $u(t_w)$ over $u(t_l)$ in the context of images $C$ and previous trials $T$. 
Note that in constructing preference pairs, we do not explicitly compare with previous trials in $T$. The comparisons we use are \emph{local} and between different utterances sampled for the same trial.

\paragraph{Alternative notions of communicative utility}
We also experiment with constructing preference pairs based on other notions of communicative utility. Specifically, we consider the effects of optimizing only for success or only for cost, rather than a combination.
In the \textsc{success} condition, we create pairs $(t_w, t_l)$ where the listener guesses correctly for $t_w$ but not $t_l$. In the \textsc{cost} condition, we create pairs $(t_w, t_l)$ where $|u(t_w)| < |u(t_l)|$, without regard to whether the listener guesses correctly.

%% file: sections/blocks/simulation-alg.tex
\begin{algorithm}[t]
    \DontPrintSemicolon
    \caption{Simulating repeated \\ reference games between models}\label{alg:sim}
    \KwIn{Speaker $S_\theta$, listener $L_\phi$, context $C$, number of trials $N$, }
    \KwOut{Samples $G$, sequence of trials $T$}
    $T \gets [\ ]$, $G \gets \emptyset$ \;
    \While{$|T| < N$}{
        {\small \tcp{Choose next target so each image appears once in a block}}
        $c \gets \texttt{NextTarget}(C, T)$ \;
        {\small \tcp{Sample $n$ speaker utterances}}
        $u_1, u_2, \ldots, u_n \sim S_\theta(\cdot|C, T, c)$ \;
        {\small \tcp{Obtain a listener guess for each utterance}}
        \For{$i = 1, 2, \ldots, n$}{
            $l_i \gets \argmax_{l} L_\phi(l|C, T, u_i)$ \;
        }
        $V \gets \{(c, u_i, l_i)|1 \leq i \leq n\}$ \; 
        $G \gets G \cup \{(C, T, V)\}$\;
        {\small \tcp{Sample a trial to continue}}
        $t_\text{next} \sim V$ \;
        {\small \tcp{Extend conversation history}}
        $T \gets T + t_\text{next}$ \;
    }
    \Return{$G$, $T$} \;
\end{algorithm}

%% file: sections/experiments.tex
\section{Experiments} \label{sec:experiments}
\subsection{Image domains} We experiment with two image domains: photographs and tangrams. We source photographs from the COCO dataset \cite{coco}. Multimodal models are trained on extensive photograph data, and have access to canonical associations between these images and words. However, in this task, models have to pick associations that are useful in the given context, enabled by conventions. We also experiment with tangrams \citep{kilogram}, which lack strong associations between images and canonical language descriptions. This makes descriptions of tangram shapes more arbitrary and challenging to interpret. However, over multiple trials, people can form conventions that coordinate on these associations and allow them to describe these images efficiently. Tangrams have been used extensively to study convention formation \cite{CLARK19861,hawkins-etal-2020-characterizing}.

\subsection{Models} We primarily use models based on Gemma-3-12B \cite{gemmateam2025gemma3technicalreport}. We use the pretrained version of the model that has not been instruction-tuned. The speaker is prompted with the context of images, the previous trials of the game, and the label for the target image in the current trial. We prompt the model with a demonstration game to show the format of game responses, but include at most one reference to an image to avoid biasing how the model handles repeated reference. For experiments with tangram images, we initially found that models were not able to generate distinct descriptions for different images. To address this, we finetuned the \emph{only the vision encoder} of the model on synthetic repeated reference games to obtain the Gemma-Tangrams-Base model. We use human-produced descriptions of tangrams collected by \citet{cogen} to create synthetic games, and use supervised finetuning to adapt the model to the visual domain. In creating data for base models, we ensure we don't use any data from real repeated interactions to avoid biasing the model to specific convention formation behavior.

\input{sections/blocks/human-human-results}

\subsection{Training} We simulate games between base speaker and listener models using \Cref{alg:sim}. We present contexts with 4 images (\Cref{sec:appendix-contexts}) and simulate $N = 20$ trials (5 blocks) for each game. For each trial, we sample $n = 4$ speaker utterances. We create preference pairs as described in \Cref{sec:method}. We then finetune the speaker with Identity Preference Optimization (IPO; \citeauthor{pmlr-v238-gheshlaghi-azar24a}, \citeyear{pmlr-v238-gheshlaghi-azar24a}). Training details are provided in \Cref{sec:appendix-training}.

\subsection{Evaluation} We evaluate our speaker models in new games with a listener, with images not seen during training. For Gemma speakers that play games over COCO images, we use greedy decoding, which we found led to the highest listener accuracy in validation games. For tangram images, while greedy decoding resulted in the highest validation accuracy, we found that it caused all models, including the Gemma-Tangrams-Base, to degenerate to repeating the same description for every repetition of an image. To mitigate this lack of diversity in the tangram speaker models, we used nucleus sampling. Additional details are provided in \Cref{sec:appendix-training}.

\input{sections/blocks/example}

\subsection{Human listener experiments}
We evaluate our speaker model in games with a human listener, and compare the performance to baseline models that play games with the same listener. As a reference, we also collect data from games between human speakers and human listeners on the same distribution of image contexts.

For experiments with a model speaker, we have human listeners play 2 games. One of these games is with our \textsc{success+cost} speaker model. The second game is with one of two baseline models. The first baseline is the \textsc{success} speaker. The second baseline for COCO games is the Gemma-3-12B instruction-tuned model, a model that has been instruction-tuned without a focus on communicative efficiency. For tangram games, we use the Gemma-Tangrams-Base model as the second baseline since it has been adapted to tangram images. 

The sequence of the two games is randomized. Each game consists of 20 trials. In each trial, the listener is presented with the context of images and the speaker's descripition, and asked to click on the image they believe the speaker to be describing. Once they guess, the listener is shown the target image, and told whether they were correct. For each trial, we also record the listener's response time -- the time between their being shown the description and their clicking on a guess. The 2 games that the listener plays are for different image contexts. At the end of the game, the listener is given some questions to answer about the game. After playing both games, the listener answers questions that compare both the speakers with which they interacted. For experiments with a human speaker, we have listeners play a single game on the same interface, and answer questions at the end of the game. 

We recruit 65 paritipants for COCO games with a model speaker, 60 participants for tangram games with a model speaker, 74 participants for COCO games with a human speaker, and 31 participants for tangram games with a human speaker. After excluding incomplete games, we are left with 126 model-as-a-speaker games for COCO and 120 for tangrams. We have 30 human-as-a-speaker games for COCO, and 12 for tangrams. Details of participant recruitment, filtering, compensation, and experiment interface are provided in \Cref{sec:appendix-prolific}.

\paragraph{Training with \textsc{success+cost} makes models communicate efficiently} We find that our \textsc{success+cost} models communicate efficiently with people, despite not being trained on any additional human-produced data. We observe increasing communicative success over repetitions of an image, while also seeing decreasing message length (\Cref{fig:human-listener-results}, left and middle). While the \textsc{success} models do achieve better accuracy, they do so with much longer messages, and don't improve over time.

We look to the listener's response time (\Cref{fig:human-listener-results}, right) to evaluate the how easy the utterances are for a person to interpret. Specifically, we consider the response time on the last repetition of an image. At the start of the game, the listener hasn't seen how the speaker describes a particular image (the convention they use), but is also not familiar with the images.
Evaluating response time on the last repetition allows us to see the effect of convention formation on a listener that has had the chance to get familiar with the context and develop convention with the speaker. We find that people who played with the \textsc{success+cost} speaker and the \textsc{success} speaker responded faster to the \textsc{success+cost} speaker (Welch's $t$-test, $p < 2 \times 10^{-5}$ for COCO and tangrams). Similarly, those who played COCO games with the \textsc{success+cost} speaker and the Gemma-3-12B-Instruct speaker responded faster to the \textsc{success+cost} speaker (Welch's $t$-test, $p < 10^{-11}$). Those who played tangram games with the \textsc{success+cost} speaker and Gemma-Tangrams-Base responded faster to the \textsc{success+cost} speaker (Welch's $t$-test, $p = 0.001$).

\input{sections/blocks/human-listener-wnr}

\input{sections/blocks/simulated-listener-results}

\paragraph{\textsc{success+cost} elicits convention formation} We find that models trained with \textsc{success+cost} communicate efficiently by forming conventions. To evaluate this, we use word novelty rate (WNR; \citeauthor{hua2024talk}, \citeyear{hua2024talk}, \Cref{sec:appendix-metrics}) which measures the rate at which new words are introduced in an utterance. It is calculated as thef word error rate between the utterance for the previous repetition of the image to the current utterance, without penalties for deletions. A low and decreasing WNR (\Cref{fig:human-listener-wnr}) suggests that the \textsc{success+cost} speaker forms conventions by reusing words from previous utterances. In contrast, we see that the other models we evaluate do not exhibit this behavior. The \textsc{success} speaker produces successful utterances, and learns that reusing words enables success, as reflected in the low WNR. However, the messages do not decrease in length, failing to exploit convention formation to be more efficient. The Gemma-3-12B-Instruct speaker produces successful utterances, but does not reduce message length and consistently innovates in its utterances, suggesting that the model doesn't form conventions. 

We find two patterns of language use by human speakers, similar to \citet{hua2025post}. Some adapt their language towards shorter, more consistent utterances, as has been found in extensive prior work. We term these ``high consistency'' speakers. Others don't adapt, choosing instead to use more verbose descriptions that continually innovate, who we term ``low consistency'' speakers. We use a similar method to \citet{hua2025post} (\Cref{sec:appendix-prolific}) to classify speakers as high consistency or low consistency. Notably, we find that human listeners are also sensitive to lack of adaptation by low consistency speakers, taking longer to respond to their utterances (\Cref{fig:human-listener-results}, right).

\input{sections/blocks/word-class}

\paragraph{Models increasingly use open class words over time} We tag utterances using the spaCy POS tagger \citep{Honnibal_spaCy_Industrial-strength_Natural_2020} and find (\Cref{fig:pos-proportion}) that the \textsc{success+cost} model's utterances contain more open class words (nouns, verbs, adjectives) over time relative to closed class words (pronouns, determiners, prepositions, etc.). Even though models aren't trained on demonstrations of human convention formation behavior, these trends are similar to what \citet{hawkins-etal-2020-characterizing} find with people.

\subsection{Simulated listener experiments}
We conduct additional experiments to evaluate speakers trained with \textsc{cost} alone, and speakers based on another pretrained model. For these experiments, we use a GPT-5-mini listener (\citet{hua2024talk} find that large multimodal models perform on par with humans as listeners at this task). In these experiments, we evaluate only listener accuracy, message length, and WNR.

We find similar trends for \textsc{success+cost} and \textsc{success} for Pixtral-based models as for Gemma-based models. Details of training for the Pixtral-based models are provided in \Cref{sec:appendix-base-model}.

\paragraph{Training models with \textsc{cost} alone is insufficient} In experiments with a simulated listener, we find that while training with \textsc{success+cost} and \textsc{success} show similar trends to experiments with human listeners, training with \textsc{cost} alone is insufficient, as it (unsurprisingly) results in 
low communicative success. Results for tangram speaker models trained with \textsc{cost} alone are in \Cref{sec:appendix-analysis}.

%% file: sections/blocks/human-human-results.tex
\begin{figure*}[t]
    \centering
    \includegraphics[width=\linewidth]{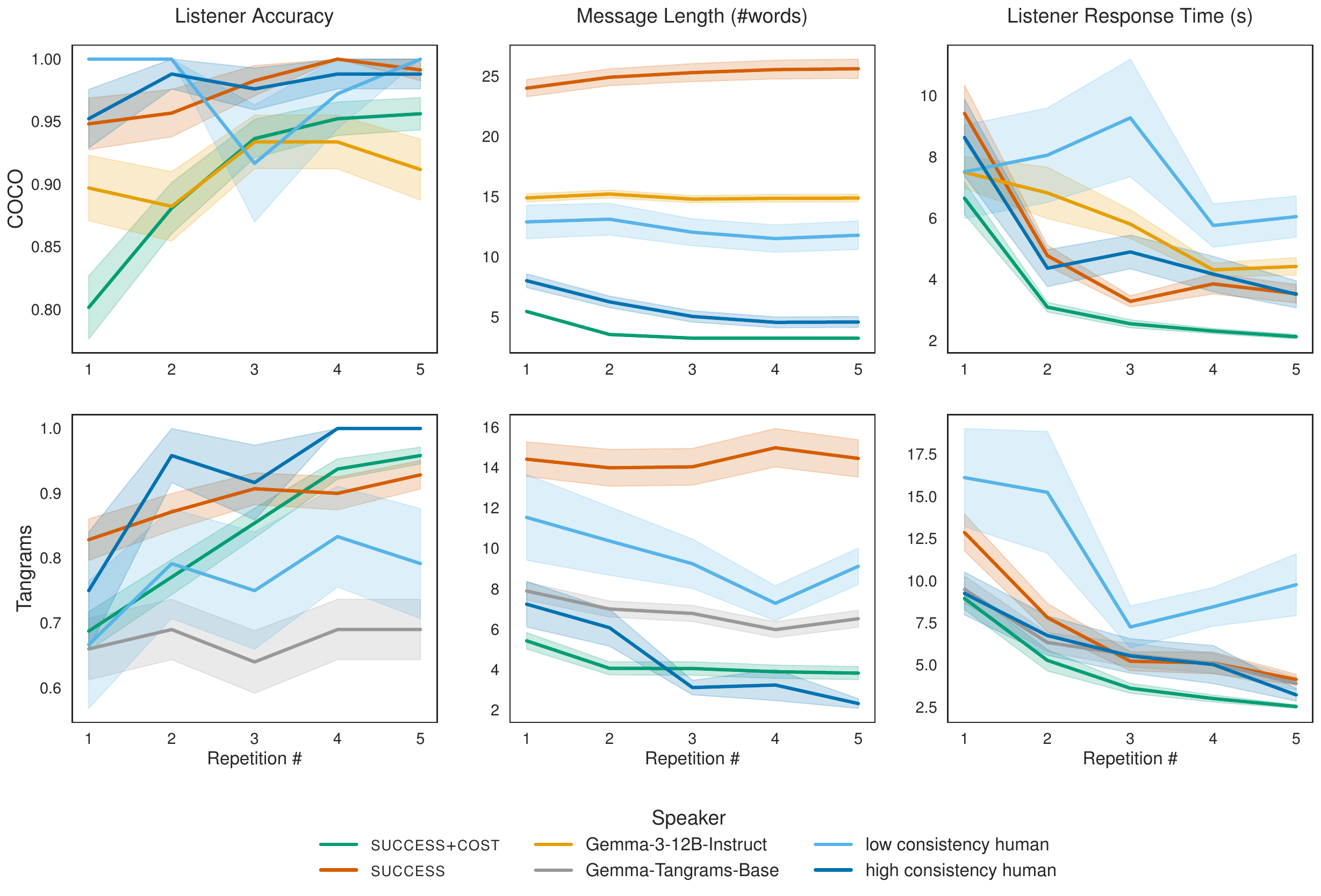}
    \caption{The \textsc{success+cost} speaker communicates increasingly successfully and efficiently over the course of the interaction with human listeners. The model achieves increasing accuracy while decreasing message length. This efficiency is also reflected in people responding faster to utterances produced by the \textsc{success+cost} speaker compared to other models. Error bars show standard error.}
    \label{fig:human-listener-results}
\end{figure*}

%% file: sections/blocks/example.tex
\begin{figure*}[t]
    \centering
    \begin{subfigure}[c]{0.45\textwidth}
        \centering
        \includegraphics[width=\linewidth]{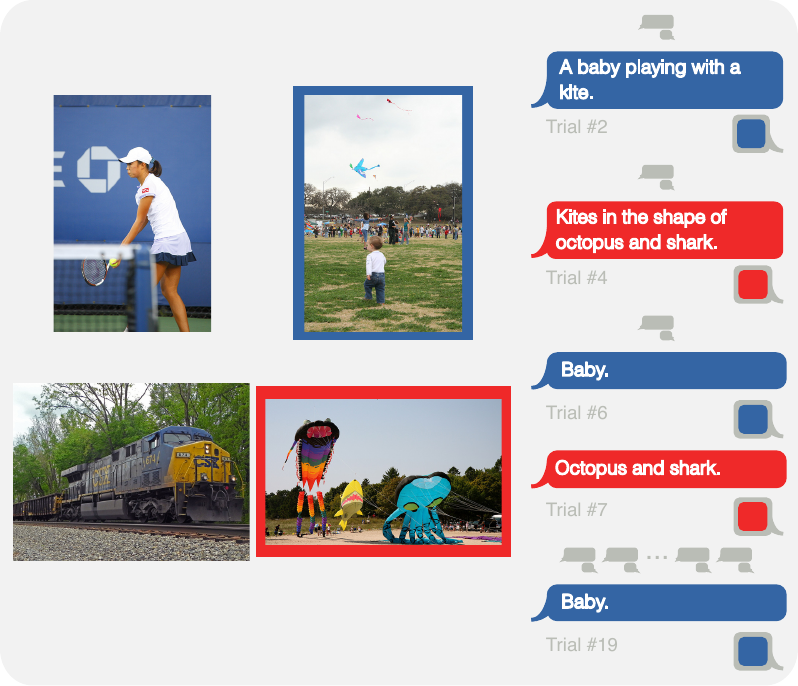}
    \end{subfigure}
    \begin{subfigure}[c]{0.45\textwidth}
        \centering
        \includegraphics[width=\linewidth]{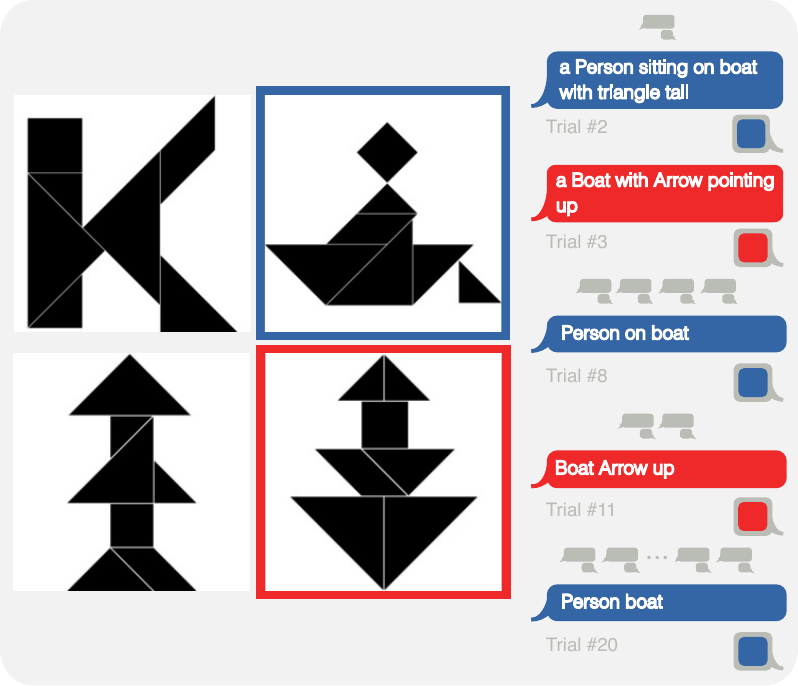}
    \end{subfigure}
    \caption{Examples of adaptation over the course of a game by the \textsc{success+cost} speaker model. In the COCO example, we see an example of the model converge to a convention that doesn't reference `kites', since it doesn't distinguish the two images. In the tangrams example, we see how the model uses a detailed description (\emph{``with triangle tail''}) in an initial turn, but in later turns drops the detail without losing success.}
    \label{fig:human-listener-example}
\end{figure*}

%% file: sections/blocks/human-listener-wnr.tex
\begin{figure}[!h]
    \centering
    \includegraphics[width=0.9\linewidth]{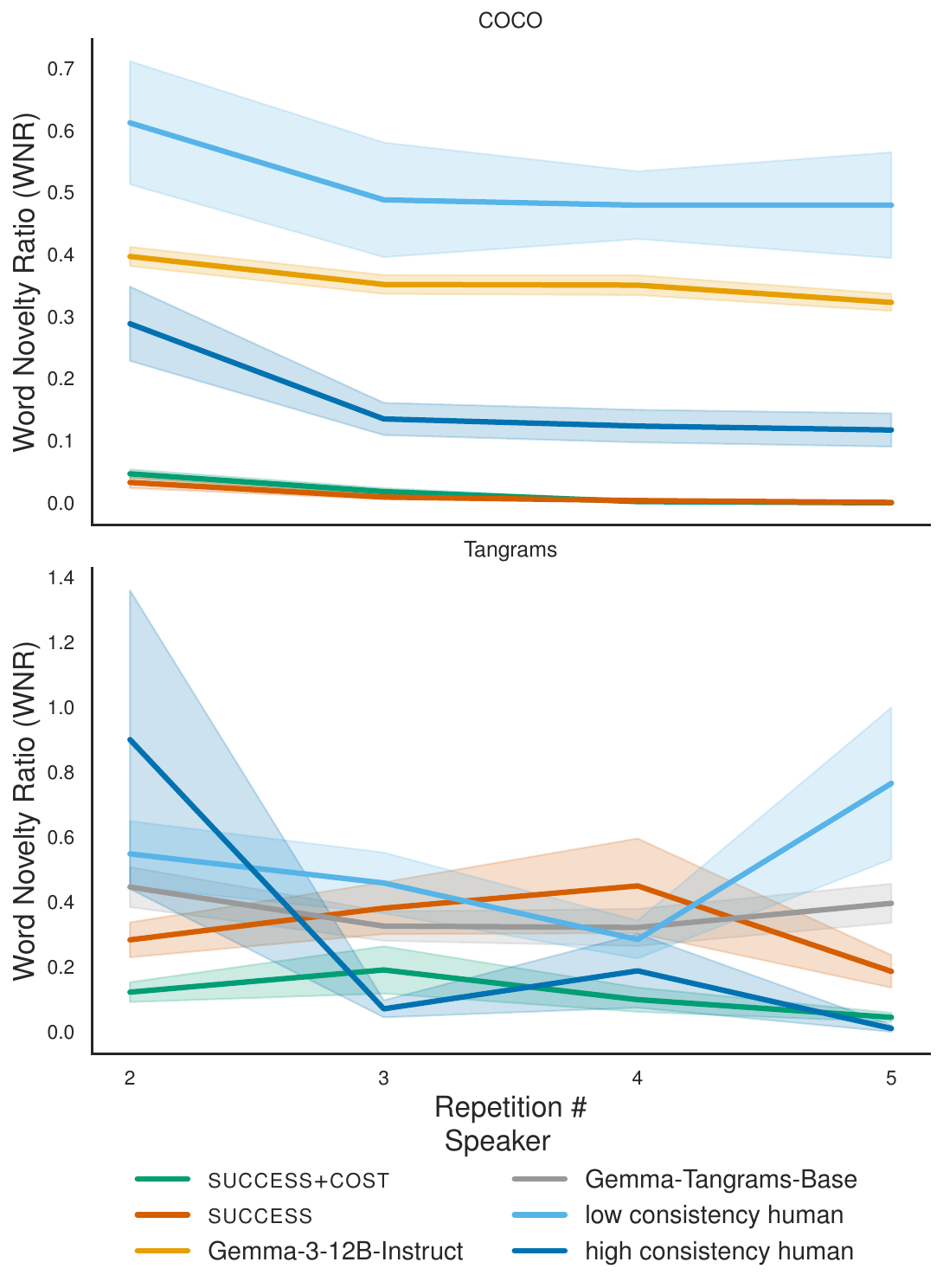}
    \caption{\textsc{success+cost} speakers achieve low and decreasing WNR. Along with decreasing message length, this suggests convention formation. \textsc{success} models achieve low WNR, but repeat entire messages without growing more efficient. Error bars show standard error.}
    \label{fig:human-listener-wnr}
    \vspace{-1em}
\end{figure}

%% file: sections/blocks/simulated-listener-results.tex
\begin{figure*}[t]
    \centering
    \includegraphics[width=\linewidth]{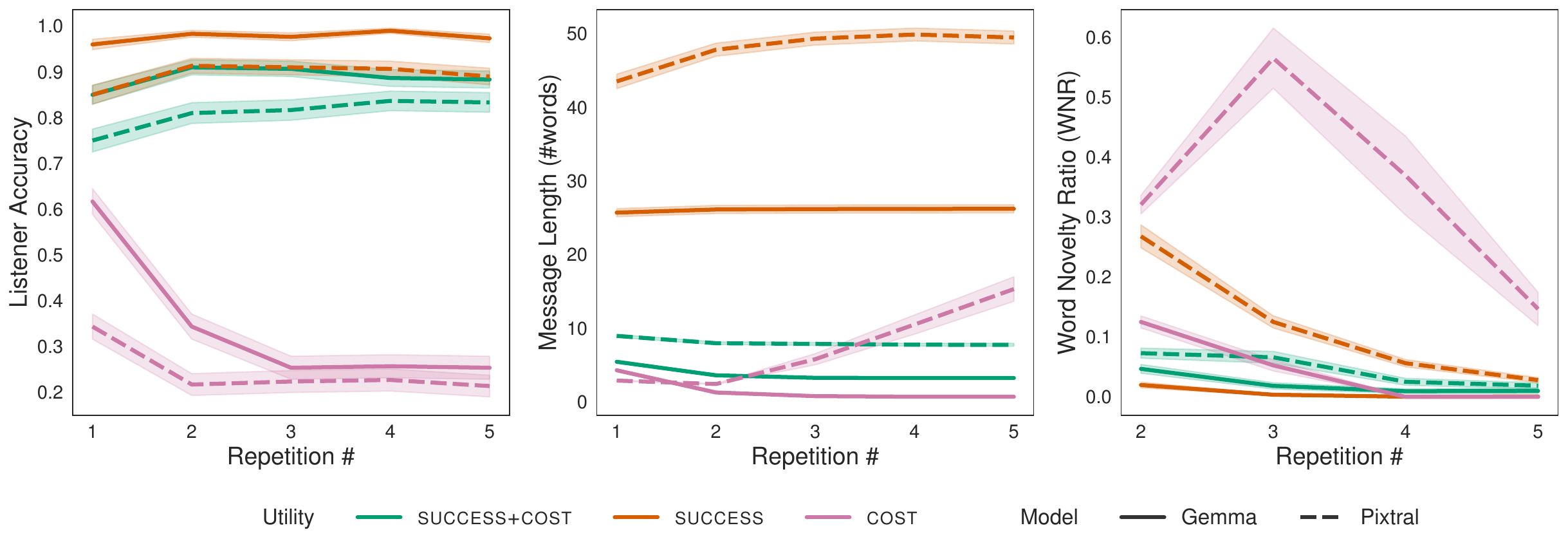}
    \caption{In COCO experiments with a GPT-5-mini listener, the \textsc{success+cost} speaker shows similar trends to human listener experiments for Gemma and Pixtral-based models. For Pixtral models, message length decreases by 13.4\% (8.95 $\to$ 7.74) by repetition 4. When training with \textsc{cost} alone, listener accuracy drops to near chance levels.}
    \label{fig:simulated-listener-results}
\end{figure*}

%% file: sections/blocks/word-class.tex
\begin{figure}[h]
    \centering
    \includegraphics[width=\linewidth]{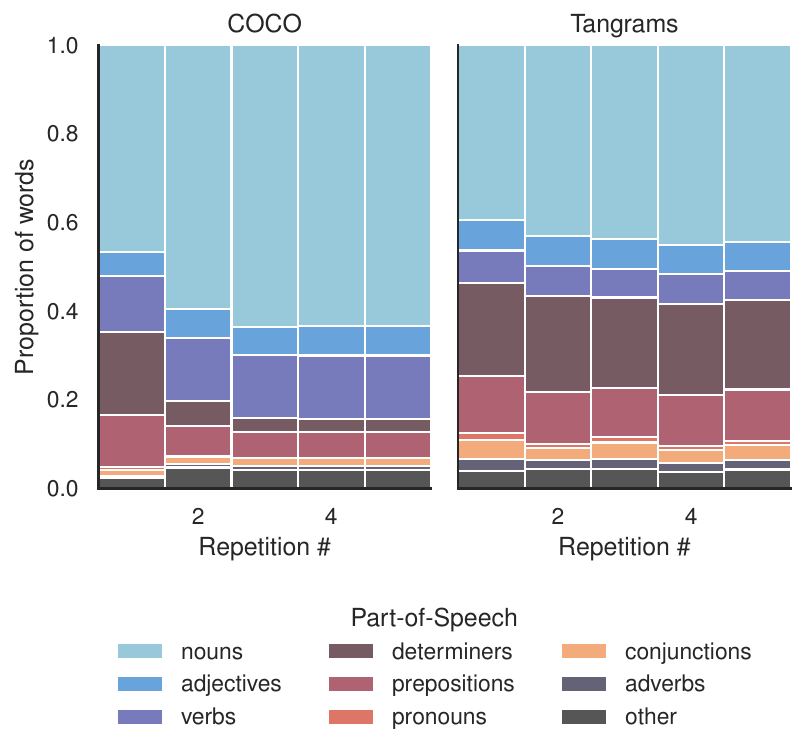}
    \caption{The \textsc{success+cost} model uses more open class words (nouns, verbs, and adjectives; in cooler hues) relative to closed class words (pronouns, determiners, prepositions, etc.; in warmer hues) in its utterances over time by preferentially dropping function words.}
    \label{fig:pos-proportion}
    \vspace{-1em}
\end{figure}

%% file: sections/discussion.tex
\section{Discussion} \label{sec:discussion}
In this paper we propose a method to train models that adapt their language to communicate more efficiently over time in repeated interactions.
We apply our method to training models that play repeated reference games over images. We find that our method elicits linguistic conventions that are legible to human listeners, even though models receive no explicit demonstrations of human convention formation behavior during training. The use of convention by the models we train also leads to faster responses from human listeners compared to models that have not been trained for communicative efficiency.
This highlights how training to communicate efficiently can improve the ability of people to respond to what a model generates.

The rewards we use to elicit this behavior are broadly applicable. We rely on general notions of communicative success (whether the listener recovered speaker's intent) and cost (the number of words). In future work, these rewards might be instantiated in other settings to allow agents to communicate more efficiently with people when carrying out complex tasks.

%% file: sections/appendix/base-model.tex
\section{Base models} \label{sec:appendix-base-model}
\subsection{COCO base models}
\paragraph{Speaker model}
For base models in COCO games, we use the pretrained versions of Gemma-3-12B \citep{gemmateam2025gemma3technicalreport} and Pixtral-12B \citep{agrawal2024pixtral12b}. We prompt these models to with one demonstration game, and the context of the game so far. For each game, we present the images once, and then reference the images using a label. So, we show the model images as \texttt{Image A: <image> Image B: <image>}, and then prompt the model to produce a description of an image by referencing its label. For the demonstration game, we choose a set of 4 COCO training images to create a context, and choose corresponding captions from the COCO dataset to use as utterances for these images. We note that these descriptions are written by people who do not have access to the other images in the context, nor to any prior conversation turns about that image, ensuring they do not reflect properties of convention formation. These captions serve as a guide for the model about the kind of descriptions we want it to produce -- without it, the model produces very long utterances that sometimes do not reference the content of the image. For the demonstration game, we also use different labels (\texttt{M}, \texttt{N}, \texttt{O}, \texttt{P}) from the main game (\texttt{A}, \texttt{B}, \texttt{C}, \texttt{D}) to ensure the model does not confuse which image is being referenced.

In addition to these steps, we had to take additional measures to ensure the Pixtral model produced utterances of appropriate length. With just the measures above, we still found the Pixtral model producing excessively long descriptions, and failing to generate an end-of-sequence token that stopped generation. As a workaround, we used a different end-of-sequence marker (\texttt{<EOM>}) which we appended to the demonstration utterances, and used as a stop string while decoding from the model. This did not correspond to a special token in the model's vocabulary, and was tokenized as a string that was part of the template to generate prompts. We found that the model was able to learn to use this token in context, and produced utterances of appropriate length.

\paragraph{Listener model} For the speaker model, we presented the images once, and referred to them only using labels. If we did the same for the listener model, it would be able to simply answer with the label associated with an image in a previous trial, without reasoning about the image at all. To ensure the listener cannot exploit associations between labels and descriptions from previous trials, we present the images to the listener twice. The second presentation is before the most recent trial presented to the listener (for which the model has to make a guess), and includes the images in a shuffled order. In initial experiments we found that shuffling the images on second presentation led to poor listener accuracy. To address this, we finetuned the listener on games in this format.

We trained a Gemma-3-12B base model to serve as a listener. Since we finetuned the model, we did not include a demonstration game. We finetuned the model to predict the label of an image based on the context of the game so far, including the utterance for the most recent trial. We construct synthetic games to train the listener using COCO descriptions. We randomly sample a set of COCO training images to create a context, and then sample a sequence of trials with that context. In each trial, we sample one of the images, a corresponding caption from the COCO dataset, and train the model to predict the label of the image given the prompt. Again, we note that since these are descriptions from the COCO dataset, they are written out of context and do not reflect properties of convention formation. 

We synthesize 250 games (5000 trials) for finetuning. We finetune all linear layers of the model with LoRA \citep{hu2022lora} with $r=16$ and $\alpha=32$. We use a batch size of 2 with 8 gradient accumulation steps (effective batch size of 16). We finetuned the model for a maximum of 5 epochs, and chose $10^{-4}$ as the learning rate warmed up over 1\% of training and linearly decayed over the remaining steps, choosing between $10^{-4}$ and $10^{-5}$ based on validation loss  on a held out set of 10 games. We choose the checkpoint with the lowest validation loss among checkpoints after each epoch. We use the \texttt{SFTTrainer} in the HuggingFace TRL library for training \citep{vonwerra2022trl}.

\subsection{Tangram base models}
\paragraph{Speaker model} For tangrams, we finetune the vision encoder of the Gemma-3-12B base model. We do this to avoid modifying the in-context learning abilities of the base model, while still adapting it to the new visual domain. We construct synthetic games for images drawn from the pretrain set (a subset of the training set; \Cref{sec:appendix-contexts}) of Kilogram images. For descriptions, we use descriptions from human-human reference games over tangram reference games collected by \citet{cogen}. They conduct multiple rounds of games, and release multiple descriptions of each Kilogram image, allowing us to create a dataset to train a model in the same way as for COCO images. Again, we note that even though these descriptions are from reference games, we use them out of their original context (we have 4 image reference games, while \citet{cogen} have 10 image reference games), and are single-turn interactions, ensuring they do not reflect properties of convention formation. Additionally, for the speaker, we sample the listener's guess at random (so the speaker isn't biased to observing only correct guesses from the listener).

We finetune the full vision encoder of a Gemma-3-12B base model on 500 games (10000 trials). We use a batch size of 1/GPU on 4 GPUs, with 4 gradient accumulation steps (effective batch size of 16). We finetuned the model for a maximum of 10 epochs, with a learning rate of $10^{-5}$ warmed up over 1\% of training and linearly decayed over the remaining steps. We choose the checkpoint with the lowest loss on a validation set of 25 games among checkpoints after each epoch.

\paragraph{Listener model} For the listener model, we finetune the vision encoder first (like for the speaker model), and then finetune all the linear layers (with LoRA) to adapt the model to image shuffling (which requires more than adapting to the visual domain). We first finetune the vision encoder on 250 games (5000 trials) generated in a similar way as for training the COCO listener model. At this stage, we don't shuffle the images on the second presentation and maintain the same order. We use a batch size of 1/GPU on 4 GPUs, with 4 gradient accumulation steps (effective batch size of 16). We finetuned the model for a maximum of 10 epochs, with a learning rate of $10^{-5}$ warmed up over 1\% of training and linearly decayed over the remaining steps. We choose the checkpoint with the lowest loss on a validation set of 25 games among checkpoints after each epoch.

Then, we use the same games to finetune all the linear layers with LoRA, using $r=16$ and $\alpha=32$. We use a batch size of 1/GPU on 4 GPUs, with 4 gradient accumulation steps (effective batch size of 16). We finetuned the model for a maximum of 10 epochs, with a learning rate of $10^{-5}$ warmed up over 1\% of training and linearly decayed over the remaining steps. We choose the checkpoint with the lowest loss on a validation set of 25 games among checkpoints after each epoch.

%% file: sections/appendix/contexts.tex
\section{Sampling contexts of images} \label{sec:appendix-contexts}
We construct image contexts based on image similarity. The effects of convention formation are known to be context-dependent \citep{Hawkins2023-HAWFPT-2}, so the choice of contexts is important. Images that are more similar are hard to discriminate in the same context, while dissimilar images are easier to tell apart. We determine image similarity using pretrained CLIP-style visual encoders.

For COCO, we create contexts of images from the COCO 2014 train set as the image bank. We embed all images using the \texttt{siglip2-base-patch16-512} \citep{tschannen2025siglip2multilingualvisionlanguage} image encoder. For each context, we sample a seed image at random, and compute the cosine similarity of that image's embedding to all images in the bank (except itself). We then divide the similarities by a temperature $\tau$ and apply the softmax function to get a categorical distribution over images. We then sample the other images without replacement from this distribution. We use $0.01$, $0.02$, $0.03$, $0.04$, and $0.05$ as temperatures for sampling contexts, with $0.01$ being the hardest contexts (most similar images), and $0.05$ being the easiest (least similar images). We sample 100 contexts at each temperature for training. For evaluation we apply the same procedure to the COCO validation set. We sample 50 contexts (10 for each temperature) for validation, and another 75 contexts for testing (15 for each temperature).

For tangrams, we use the original splits from the Kilogram dataset \citep{kilogram}. We further partition the training set into two, with one part used for pretraining (finetuning the speaker's vision encoder and the listener), and the other part used for simulating games to produce preference training data. We use the CLIP encoders released by \citet{kilogram} (for whole, black images, following \citeauthor{cogen}, \citeyear{cogen}) to compute image similarity, and sample contexts at temperature $0.125$, $0.25$, $0.5$, and $1.0$. We sample 100 contexts at each temperature for training. We also apply the same procedure to the validation split of Kilogram images to construct the validation set (10 contexts at each temperature). For the test set, we use the test split of Kilogram images, and sample 40 contexts each at temperatures $0.125$, $0.25$.

%% file: sections/appendix/training.tex
\section{Preference optimization} \label{sec:appendix-training}
For COCO we simulate 500 games (10000 trials), and construct preference optimization pairs using the criteria described in \Cref{sec:method}. For tangrams, we simulate 400 games (8000 trials). We use top-$p$ sampling with decoding temperature $1.0$ and $p=0.95$ to sample utterances. Given that the preference pairs are constructed from samples drawn from the model, the exact number of preference pairs varies based on the criteria used to construct them. We use a fixed number of games, and use all the preference pairs constructed from these games to train the model. During preference optimization, we use the same demonstration game used while simulating games.

We adapt all linear layers of the model the model with LoRA adapters ($r=32$ and $\alpha=32$), using IPO as the preference optimization algorithm. We train for 3 epochs. We use a batch size of 1/GPU on 4 GPUs, with 8 gradient accumulation steps (effective batch size of 32). We choose a learning rate from $10^{-6}$ and $10^{-5}$ and a KL regularization parameter $\beta$ between $0.1$, $0.2$, $0.3$ and $0.4$ using a validation procedure (described below). We choose $10^{-5}$ as the learning rate and $\beta=0.3$ for the Gemma model on COCO images and use the same parameters for all other Gemma experiments (COCO and tangram models, all utility functions). For Pixtral models (all utility functions), we use a learning rate of $10^{-6}$ and $\beta=0.3$, without searching over different values of these parameters. For evaluation of COCO models, we sweep over decoding temperature $0$ (greedy decoding), $0.3$, $0.7$ and $1.0$. Based on the validation procedure, we chose greedy decoding for Gemma and $0.7$ for Pixtral.

For validation, we construct a set of images contexts from a held out set of images. We simulate games for context, generating 4 utterances for each trial (like we do when simulating games for training). We then use average listener accuracy across all samples at all trials (this can be viewed as the area under the curve of accuracy over repetitions of images) as the validation criterion.

%% file: sections/appendix/metrics.tex
\section{Metrics} \label{sec:appendix-metrics}
To evaluate message length, we use the spaCy package (\texttt{en\_core\_web\_sm}, \cite{Honnibal_spaCy_Industrial-strength_Natural_2020}) to tokenize the utterance and count all the tokens. For WNR, we first compute the word novelty distance (WND). We follow \cite{hua2024talk} in selecting the content words (with POS tag \texttt{NOUN}, \texttt{ADJ}, \texttt{VERB}, \texttt{ADV}, \texttt{PROPN}, \texttt{NUM}, \texttt{PRON}, or \texttt{ADP}), lemmatizing them with spaCy, and computing the edit distance between the previous utterance used to describe an image and the current utterance with no penalties for deletions. We then compute the WNR as the WND divided by the length of the previous utterance.

We also map POS tags to part-of-speech types (for \Cref{fig:pos-proportion}) as follows:
\begin{table}[h]
\centering
\begin{tabular}{ll}
\toprule
\textbf{POS Tag} & \textbf{Part-of-speech type} \\
\midrule
ADJ    & adjectives    \\
ADP    & prepositions  \\
ADV    & adverbs       \\
AUX    & other         \\
CCONJ  & conjunctions  \\
DET    & determiners   \\
INTJ   & other         \\
NOUN   & nouns         \\
NUM    & other         \\
PART   & other         \\
PRON   & pronouns      \\
PROPN  & nouns         \\
PUNCT  & other         \\
SYM    & other         \\
VERB   & verbs         \\
X      & other         \\
SCONJ  & conjunctions  \\
\bottomrule
\end{tabular}
\caption{Mapping from POS tags to part-of-speech types used in \Cref{fig:pos-proportion}.}
\label{tab:pos-mapping}
\end{table}

%% file: sections/appendix/prolific.tex
\section{Human study} \label{sec:appendix-prolific}
\begin{table}[t]
    \centering
    \begin{tabular}{llr}
    \toprule
         &  & \#\\
         & Speaker &  \\
        \midrule
        \multirow[t]{5}{*}{COCO} & \textsc{success+cost} & 63 \\
         & Gemma-3-12B-Instruct & 34 \\
         & \textsc{success} & 29 \\
         & Human & 30 \\
    \cline{1-3}
        \multirow[t]{5}{*}{Tangrams} & \textsc{success+cost} & 60 \\
         & Gemma-Tangrams-Base & 25 \\
         & \textsc{success} & 35 \\
         & Human & 12 \\
    \bottomrule
    \end{tabular}
    \caption{The number of games with each speaker.}
    \label{tab:human_study_games}
\end{table}

We recruit participants through Prolific. Based on pilot experiments, we found that two model speaker--human listener games take 15 minutes to complete, and one human speaker--human listener game takes 15 minutes to complete (human speakers take longer to type a message than a model takes to produce one). Based on these estimates, we compensated participants with US\$1.50 for each model speaker-human listener game, and US\$3.00 for each human speaker--human listener game. This amounts to an hourly rate of US\$12.00. We also incentivized participants with a US\$0.04 bonus for every correct trial (for a listener, this was choosing the correct image, for a speaker, this was producing an utterance that led to the listener choosing the correct image). We recruit participants who are located in the United States, are fluent in English, have at least 1000 approved Prolific submissions, and an approval rate of at least 99\% on Prolific. Our study was deemed exempt from review by our Institutional Review Board.

We conducted experiments separately for model-as-a-speaker games and human-as-a-speaker games. For the model-as-a-speaker setting, the games proceeded as follows:
\begin{itemize}
    \item The participant signs a consent form (agreeing to their data being used for this research study, and declaring that they are at least 18 years of age, and are located in the US), and read instructions for the task.
    \item The participant proceeds to play a game with the first model. This model is chosen by deciding with probability $0.5$ whether it will be \textsc{success+cost} model or one of the baseline models. If it is one of the baseline models, one of the two baseline models is chosen with probability $0.5$.
    \item The participant answers questions about the first game they played.
    \item The participant proceeds to play a game with the second model. If the \textsc{success+cost} model was chosen in the first game, the second model one of the baseline models chosen with probability $0.5$. If it is one of the baseline models, then the second model is the \textsc{success+cost} model.
    \item The participant answers questions about the second game they played.
    \item The participant is given a set of questions to answer comparing both the models they interacted with.
\end{itemize}
For the human-as-a-speaker setting, participants signed a consent form, read instructions for the task, played a game, and then answered questions about the game if they were a listener.

For model-as-a-speaker games, we excluded all data from participants who did not complete both games. For the human-as-a-speaker setting, we excluded incomplete games (due to issues such as disconnection). A bug in our human-human study interface also led to some participants re-entering the pool and getting paired to play another game. We excluded all games where either participant was not playing their first game.
The number of games with each speaker is shown in \Cref{tab:human_study_games}. For tangrams, we also excluded data from one human speaker--human listener pair where the human speaker did not send informative messages, leading to the listener getting chance accuracy.

\subsection{Interface} 

\begin{figure*}[t]
    \centering
    \includegraphics[width=0.95\textwidth]{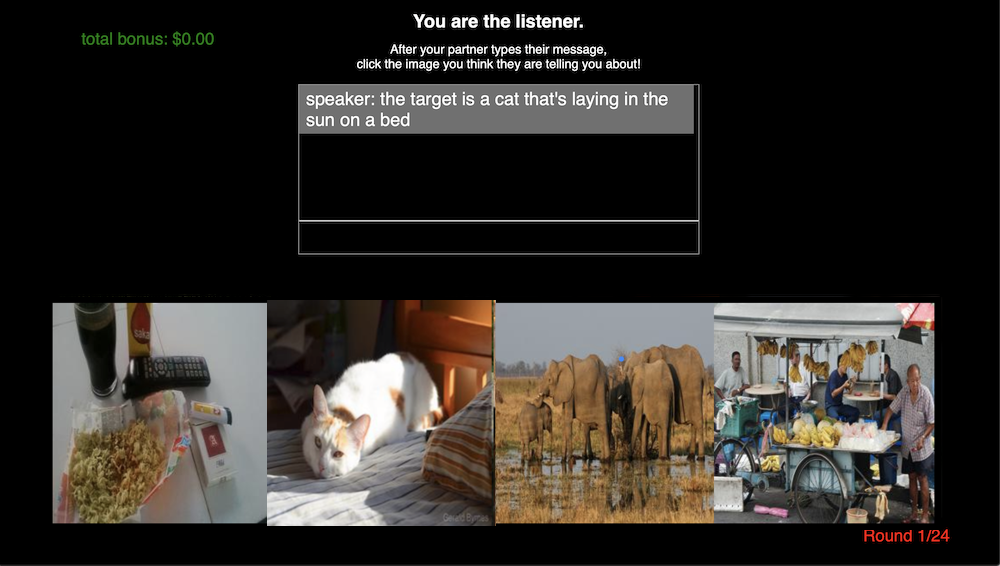}
    \caption{The listener's interface. While the speaker is typing, or the model is generating a response, the box indicates that the other player is typing. Once their message is sent, it is displayed in the box.}
    \label{fig:listener-interface}
\end{figure*}

\begin{figure*}[t]
    \centering
    \includegraphics[width=0.95\textwidth]{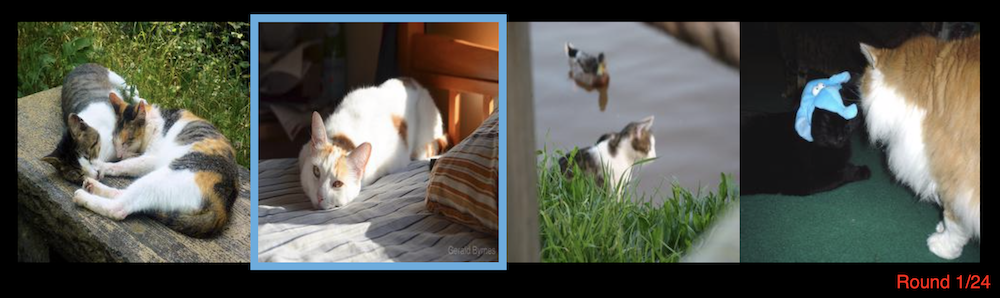}
    \caption{The speaker's view of images. The highlighted image is the one that the speaker is describing.}
    \label{fig:speaker-interface}
\end{figure*}

We conducted our study on a web interface, adapted from the one released by \citet{hawkins-etal-2020-continual}. For model-as-a-speaker games, we used colors to refer to the models, so as to not reveal their identity/nature to the participant. We always referred to the \textsc{success+cost} model as the blue robot, and the \textsc{success} model as the yellow robot, and the Gemma-Tangrams-Base/Gemma-3-12B-Instruct model as the orange robot.

The listener interface is shown in \Cref{fig:listener-interface}. In addition to this interface, we recorded the participant's response time for each trial, ensuring to keep the clock running only while the participant's tab was active. The images are shuffled after each trial.

For human-as-a-speaker games, we used the same interface as the model-as-a-speaker games for the listener. The speaker is shown the images as in \Cref{fig:speaker-interface}, and are instructed to describe the highlighted image.

\subsection{Questions}
\begin{figure*}[t]
    \centering
    \includegraphics[width=0.95\textwidth]{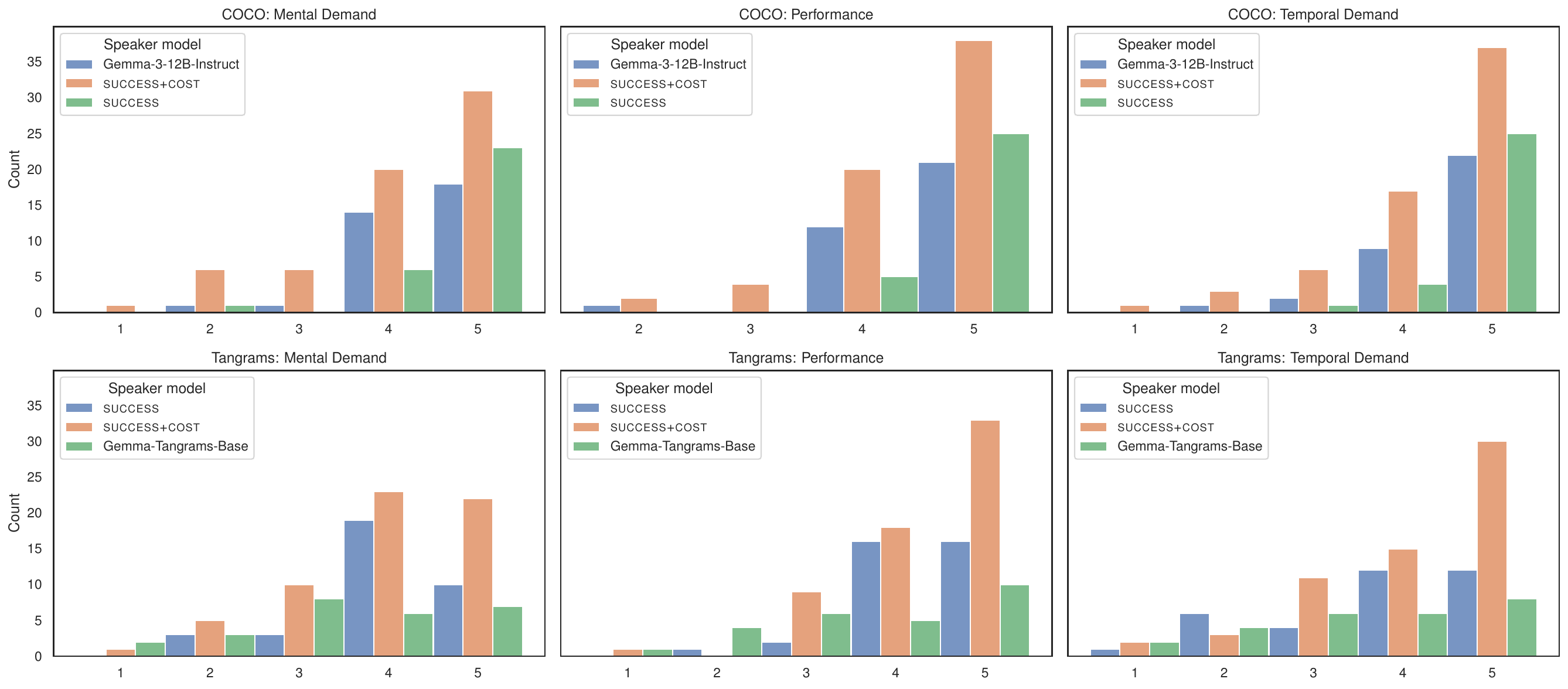}
    \caption{Questions asked to participants about the model with which they interacted.}
    \label{fig:model-survey}
\end{figure*}

\begin{figure*}[t]
    \centering
    \includegraphics[width=0.95\textwidth]{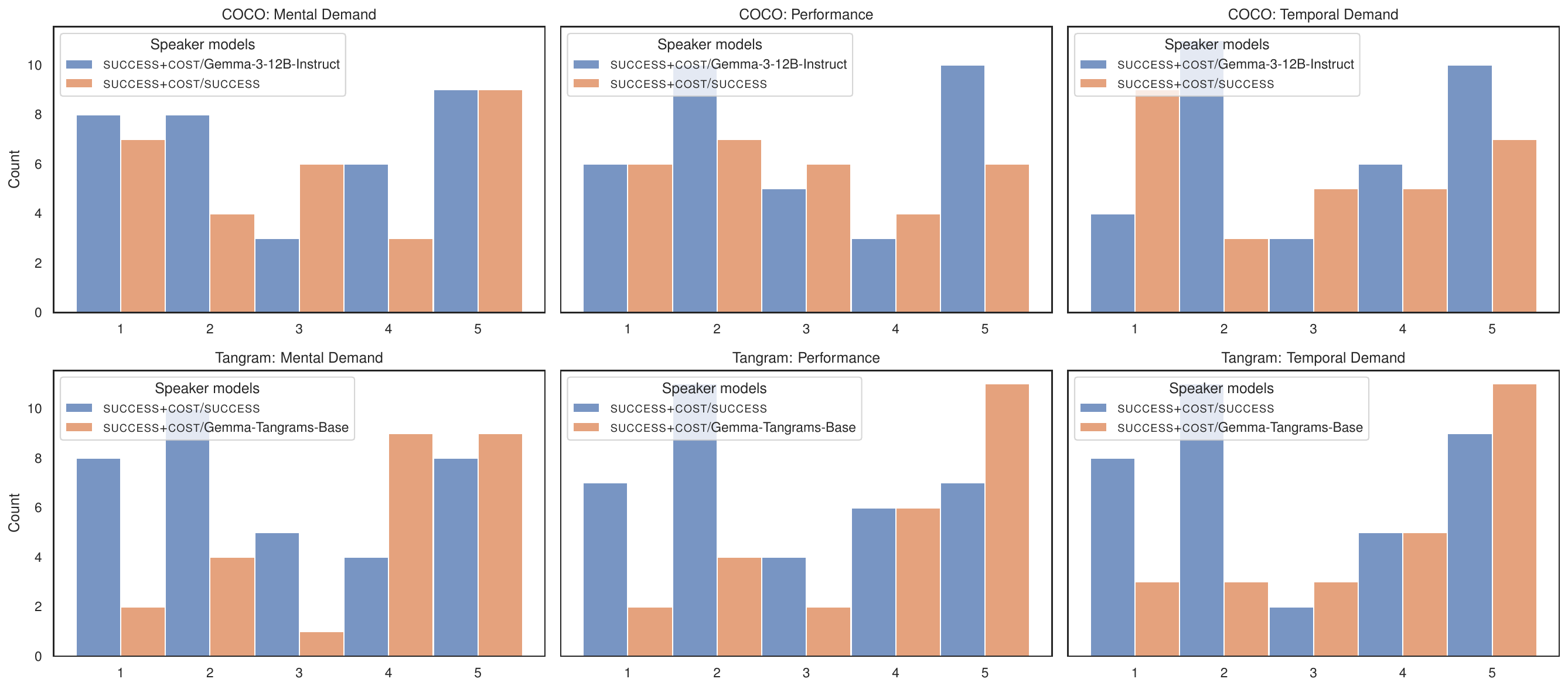}
    \caption{Questions asked to participants comparing different models.}
    \label{fig:comparative-survey}
\end{figure*}

We asked participants questions at different stages, each time after they had completed a game. After playing a game with a model, they answered the following questions, asking whether they agreed with the following statements on a 5 point scale from 1 (strongly disagree) to 5 (strongly agree).
\begin{itemize}
    \item \textbf{Mental Demand}: It was easy to understand the speaker’s descriptions.
    \item \textbf{Temporal Demand}: I was able to quickly understand the speaker’s descriptions.
    \item \textbf{Performance}: I was able to successfully interpret the speaker’s descriptions.
\end{itemize}

After completing both games, they answered the following questions, phrased based on whether they played with the yellow or orange model. The responses are on a 5 point scale from 1 (strongly disagree) to 5 (strongly agree).
\begin{itemize}
    \item \textbf{Mental Demand}: I found it easier to understand the blue robot's descriptions than the (orange/yellow) robot's descriptions.
    \item \textbf{Temporal Demand}: I was able to more quickly understand the blue robot's descriptions than the (orange/yellow) robot's descriptions.
    \item \textbf{Performance}: I was more successful at interpreting the blue robot's descriptions than the (orange/yellow) robot's descriptions.
\end{itemize}

For COCO games, people generally gave the models high ratings on all parameters (\Cref{fig:model-survey}, top row), and did not have any clear preferences between the models (\Cref{fig:comparative-survey}, top row). In tangram games, people showed a preference for the \textsc{success+cost} model over Gemma-Tangrams-Base (\Cref{fig:comparative-survey}, bottom row), and generally had lower ratings for the Gemma-Tangrams-Base model (\Cref{fig:model-survey}, bottom row).

\subsection{Characterizing the consistency of human speakers}
Like \citet{hua2025post}, we found that not all human speakers showed the kind of adaptation behavior observed in previous work. While the reasons for this are not known, we follow \citet{hua2025post} in dividing speakers into two groups: high consistency and low consistency by clustering speakers based on word novelty. For each human speaker, we compute the word novelty distance (WND; which is the WNR without normalizing by the length of the previous utterance) in the last two blocks of the game. By that point in the game, we expect stable conventions to have formed. We see that not all speakers have low WND at this stage, with some maintaining a very high WND even late into the game.

We follow \citet{hua2025post} in fitting a Gaussian mixture model (GMM) to the average WND in the last two blocks of the game. For COCO games, we speaker over 1--8 components, and 25 random seeds to find a model with the lowest Bayesian Information Criterion (BIC). The best model has 7 components, with means [0.134, 1.187, 2.655, 5.625, 9.875, 4.625, 3.750]. The first two clusters have low average WND, indicating high consistency speakers. The remaining clusters have high average WND, indicating low consistency speakers. We have 21 high consistency speakers and 9 low consistency speakers.

For tangram games, given that we have fewer games, we searched over 1--4 components and 25 random seeds. The best model has 4 components, with means [0.417, 2.667, 5.417, 3.750]. The first cluster has low average WND, indicating high consistency speakers. The remaining clusters have high average WND, indicating low consistency speakers. We are left with 6 low consistency speakers and 6 high consistency speakers.

%% file: sections/appendix/analysis.tex
\section{Tangrams model-as-a-listener results} \label{sec:appendix-analysis}

\begin{figure*}[t]
    \centering
    \includegraphics[width=\linewidth]{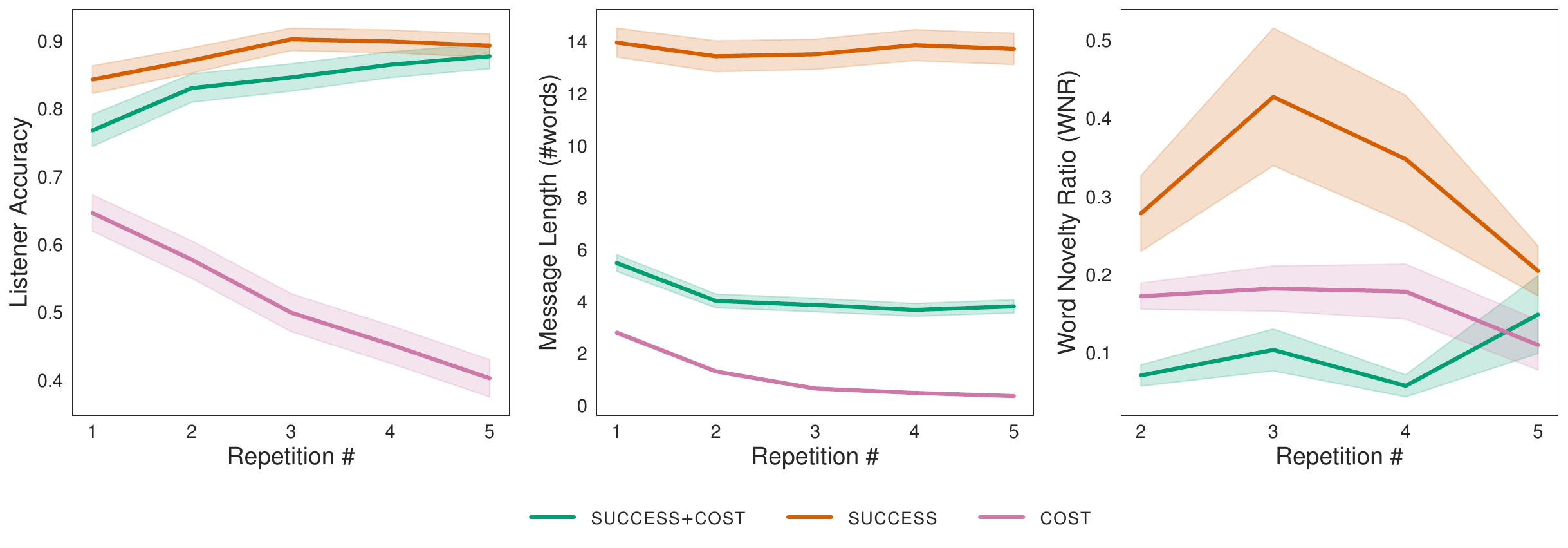}
    \caption{In tangram experiments with a model listener, the \textsc{success+cost} and \textsc{success} speakers show similar trends to human listener experiments. When training with \textsc{cost} alone, we see that the listener accuracy decreases, even though the messages are shorter.}
    \label{fig:tangrams-model-as-listener-results}
\end{figure*}

We conduct experiments on tangram images with a model listener. To ensure we have a strong model for the visual domain, we use the same listener model that we use for generating feedback during training. We perform inference the same was as for human listener experiments. Results are shown in \Cref{fig:tangrams-model-as-listener-results}, and reflect similar trends to human listener experiments, and to COCO games for the \textsc{cost} speaker.